\documentclass[sigconf]{aamas}  

\usepackage{booktabs}

\usepackage{flushend}
\setcopyright{ifaamas}  
\acmDOI{doi}  
\acmISBN{}  
\acmConference[AAMAS'20]{This work was accepted in the 11th International Workshop on Optimization and Learning in Multiagent Systems, in the International Conference on Autonomous Agents and Multi-Agent Systems 2020}{May 2020}{Auckland, New Zealand}  
\acmYear{2020}  
\copyrightyear{2020}  
\acmPrice{}  

\usepackage{xcolor}

\DeclareMathOperator*{\argmax}{arg\,max}

\newcommand{\ssim}[1]{\mathrm{sim}(#1)}
\newcommand{\cvg}[1]{\mathrm{cvg}(#1)}
\newcommand{\cp}[1]{\mathrm{cp}\big(#1\big)}

\newcommand{\np}{\mathcal{NP}}

\usepackage[ruled,linesnumbered]{algorithm2e}
\SetKwFor{For}{for}{do}{}
\SetKwFor{While}{while}{do}{}
\SetKwIF{If}{ElseIf}{Else}{if}{:}{elif :}{else :}{}
\SetKwInOut{Input}{input}\SetKwInOut{Output}{output}
\usepackage{url}
\usepackage{bbm}
\usepackage{subcaption}
\usepackage{todonotes}

\begin{document}

\title{TAIP: an anytime algorithm for allocating student teams to internship programs}  
\titlenote{Research supported by projects AI4EU (H2020-825619), Wenet (H2020-FETPROACT-2018-01), LOGISTAR (H2020-769142), and 2019DI17.}

\author{Athina Georgara}
\affiliation{%
  \institution{Artificial Intelligence Research Institute (IIIA-CSIC)}
}
\email{ageorg@iiia.csic.es}
\author{Carles Sierra}
\affiliation{%
  \institution{Artificial Intelligence Research Institute (IIIA-CSIC)}
}
\email{sierra@iiia.csic.es}
\author{Juan A. Rodriguez-Aguilar}
\affiliation{%
  \institution{Artificial Intelligence Research Institute (IIIA-CSIC)}
}
\email{jar@iiia.csic.es}

\begin{abstract}  
In scenarios that require teamwork, we usually have at hand a variety of specific tasks, for which we need to form a team in order to carry out each one.
Here we target the problem of matching teams with tasks within the context of education, and specifically in the context of forming teams of students and allocating them to internship programs.
First we provide a formalization of the {\em Team Allocation for Internship Programs Problem}, and show the computational hardness of solving it optimally.
Thereafter, we propose {\em TAIP}, a heuristic algorithm that generates an initial team allocation which later on attempts to improve in an iterative process.
Moreover, we conduct a systematic evaluation to show that TAIP reaches optimality, and outperforms CPLEX in terms of time.
\end{abstract}


\maketitle

\section{Introduction}
In the context of education, it is increasingly common that students spend some time doing practical work in a company as part of their curriculum. This work is sometimes remunerated: companies benefit from this program as they get motivated students that will work for reduced wages, and students benefit from a first contact with the labour market. It has been found that the employability of students at the end of their studies increases thanks to these internships. Nowadays, education authorities match students with companies mostly by hand. This paper formalises this matching process as a combinatorial optimization problem, proposes some heuristic algorithms and studies their computational complexity.

Team formation with respect to skills/expertise is a well studied topic of interest within the AI and MAS community \cite{AndrejczukBRSM18}.
~\cite{inproceedings} tackle the problem of team formation considering skills, communication costs, and tasks that progressively arrive in time.
In the same direction, \cite{10.1007/978-3-642-33486-3_31} propose a heuristic algorithm for forming one team of experts for a specific task.
~\cite{10.1007/978-3-319-46882-2_2} propose several heuristic algorithms for forming a single robust team in order to compete a given set of tasks.
The authors in~\cite{56283} target the problem of partitioning a group of individuals into equal-sized teams so that each one will resolve the same task.
Here we consider the problem of allocating  individuals into teams of different sizes in order to resolve different tasks.
In fact, our problem can be viewed as a generalization of~\cite{56283}.

In this work, we present and formalise an actual-world problem,  the so-called {\em Team Allocation for Internship Programs (TAIPP)}.
We characterise the complexity of the TAIPP and the search space that an algorithm that solves it must cope with.
We propose how to encode the TAIPP as a linear program so that it can be solved by a general purpose LP solver. Furthermore, we propose a novel, anytime heuristic algorithm that exploits the structure of the TAIPP.
As we will show, our proposed algorithm outperforms the general purpose optimizer IBM CPLEX in terms of time: it always reaches the optimal solution at least 55\% faster than CPLEX, and reaches a quality of 80\% in less than 20\% of the time we need to construct the input for CPLEX.

As such, in what follows, in Sec~\ref{sec:problem} we formally describe the TAIPP, provide formal definitions of the problem's components, and study the complexity of the problem. In Sec~\ref{sec:optimization:problem} we provide the encoding for a linear program solver.
In Sec~\ref{sec:TAIP} we propose our heuristic algorithm; while in Sec~\ref{sec:experiments} we conduct a systematic evaluation and show the effectiveness of our algorithm.

\section{Problem Formalization}\label{sec:problem}
In this section we present the individual components of the problem, discuss their intuition, and provide formal definitions. 
We begin with the formalization of {\em internship programs} and {\em students}, along with a thorough discussion on the essential notion of {\em competencies}.
Then we proceed on presenting our notion of competence coverage, and show how to compute it.
\vspace{-5pt}
\subsection{Basic elements of the allocation problem}\label{sec:basic:elements}

An {\em internship program} is characterised by a set of requirements on student competencies and team size constraints. 
For instance, think of an internship program in a computer tech company: there are 4 competence requirements {\em (a)} machine learning principles, {\em(b)} coding in python, {\em (c)} web development, and {\em(d)} fluency in Spanish language, while the required team size is 3 members; as such, for this program we need a team of three students that as a team possesses the four required competencies.

In general we can have a large variety of other constraints, such as temporal or spatial constraints, i.e., when and where the internship can be realised. However, within the scope of this paper, we only focus  on team size constraints.
The required competencies are often accompanied by their level and importance.
Formally, an {\em internship program} $p$ is a tuple $\langle C,l,w,m \rangle$, where $C$ is the set of required competencies, $l:C\to\mathbb{R}^+\cup\{0\}$ is a required competence level function, $w : C\to (0,1]$ is a function that weighs the importance of competences, and $m\in \mathbb{N}_+$ is the team size required.
The set of all internship programs is denoted with $P$, with $|P| = M$.

A {\em student} is characterised by their competencies, and their competence levels.
Formally, a student $s$ is represented as a tuple $ \langle C, l\rangle$, where $C$ is the set of already acquired competencies, and $l:C\to\mathbb{R}_+\cup\{0\}$ is a competence level function, and hence $l(c)$ is the student's competence level for competence $c$.
The set of all students is denoted with $S$, with $|S|=N$.
Given $p\in P$, we denote the set of all size-compliant teams for $p$ as $\mathcal{K}_p = \{K \subseteq S : |K| = m_p\}$.\footnote{Note: we use the subscript $s$ to refer to the set of competencies, competence level function, etc. of a student $s\in S$, and the subscript $p$ to refer to the same elements of the internship's $p\in P$.}

But size is not enough, we need that the members of a team are suitable for the {\em competencies} requested by a company. We assume that there is a predefined ontology that provides a fixed and finite set of competencies $C$ along with relations among them. We further assume that the ontology is a tree graph, where children denote more specific competencies than those of their parents. Formally, an ontology is a tuple $o=\langle C,E\rangle$ with $C$ being the competencies-nodes and $E$ the edges. 
The metric over ontologies that we will use next is the  {\em semantic similarity}.
The {\em semantic similarity} is given by
\begin{equation}
    \ssim{c_1,c_2} = \begin{cases}
    1, &\text{if } l=0\\
    e^{-\lambda l} \frac{e^{\kappa h} - e^{-\kappa h}}{e^{\kappa h} + e^{-\kappa h}}, &\text{otherwise}
    \end{cases}
\end{equation}
where $l$ is the shortest path in the tree between $c_1$ and $c_2$, $h$ is the depth of the deepest competence subsuming both $c_1$ and $c_2$, and $\kappa, \lambda \in [1,2]$ are parameters regulating the influence of $l$ and $h$ to the similarity metric.
This is a variation of the metric introduced in~\cite{1209005}, which guarantees the reflexive property of similarity, that is, a node is maximally similar to itself, independently of its depth. In other words, nodes at zero distance ($l=0$) have maximum similarity.
Similarly to~\cite{Osman:2014:TMA:2542182.2542198}, the semantic similarity between two competence lies in $[0,1]$.

\subsection{Computing competence coverage for students and teams}

In order to evaluate how well a student fits with an internship we need some notion of coverage for each competence required by an internship by the actual competencies of a student. Thus, we define the {\em student coverage} of competence $c$ by a set of competencies $A\subseteq C$ as $  \cvg{c,A} = \max_{c' \in A} \{\ssim{c,c'}\}$.

And then, naturally, given  a program $p$ with required competencies $C_p$ and a student $s$ with acquired competencies $C_s$ the competence coverage of program $p$ by student $s$ is:
\begin{equation}
    \cvg{s,C_p} = \prod_{c \in C_p} \cvg{c,C_s} = \prod_{c \in C_p} \max_{c' \in C_s} \{\ssim{c,c'} \}
\end{equation}
Moving now from a single student $s \in S$ to a team of students $K \subseteq S$, we need first to solve a {\em competence assignment problem}.
That is, we need to assign to each student $s\in K$ a subset of competencies of $C_p$, and assume that student $s$ is responsible for (in charge of) their assigned competencies.
According to~\cite{56283} we have that:
\begin{definition}[{\em Competence Assignment Function (CAF)}]
Given a program $p \in P$, and a team of students $K \subseteq S$, a competence assignment $\eta_p^K$ is a function $\eta_p^K:K \to 2^{C_p}$, satisfying $C_p = \bigcup_{s \in K} \eta_p^K(s)$.
\end{definition}
The set of competence assignments functions for program $p$ and team $K$ is noted by $\Theta_p^K$.
The inverse function $\eta_p^{K\ -1}:C_p \to 2^K$ provides us with the set of students in $K$ that are assigned to competence $c\in C_p$.

However, not all competence assignments are equally accepted. For example, consider a program $p$ (with $C_p = \{c_1,c_2,c_3,c_4,c_5\}$), and a team $K=\{s_1,s_2,s_3\}$. An assignment $\eta_p^K$ such that $\eta_p^K(s_1) = C_p$ and $\eta_p^K(s_2) = \eta_p^K(s_3) = \emptyset$ seems to be unfair---assigning all competencies as student $s_1$'s responsibility---,while assignment $\tilde{\eta}_p^K$ such that 
$\tilde{\eta}_p^K(s_1) = \{c_1,c_3\}$, $\tilde{\eta}_p^K(s_2)= \{c_2,c_5\}$ and $\tilde{\eta}_p^K(s_3) = \{c_4\}$ is more fair, in terms of allocating responsibilities.
In the setting of internship programs, we prefer assignments such that all students are actively participating, i.e., assignments such that $\eta_p^K(s) \neq \emptyset$ for each student $s$ (the so-called {\em inclusive assignments} in ~\cite{ewaThesis}).
At the same time, we would prefer not to `overload' a few students with excessive responsibilities, but selecting fair competence assignments.
This is captured by the following definition:
\begin{definition}[{\em Fair Competence Assignment Function (FCAF)}]\label{def:fair_competence_assignment}\em
Given a program $p$, and a team of students $K \subseteq S$, a fair competence assignment $\eta_p^K$ is a function $\eta_p^K:K \to 2^{C_p}$, satisfying $C_p = \bigcup_{s \in K} \eta_p^K(s)$, $1\leq |\eta_p^K(s)| \leq \lceil \frac{|C_p|}{|K|} \rceil \ \forall s \in K$, and $1 \leq |\eta_p^{K\ -1}(c)| \leq \lfloor\frac{|K|}{|C_p|}\rfloor + 1$.
\end{definition}

Now, given a competence assignment $\eta_p^K$, we define the {\em competence proximity} of a student $s$ {\em wrt} a program $p$. 
To do so we take into consideration the importance of each competence and the
students coverage of the assigned competencies.
In the competence proximity we want to encode the following scenarios: 
\begin{itemize}
    \item the competence proximity should be as high as possible when the coverage of a competence by a student is maximum;
    \item the competence proximity should be as high as possible when the competence is not important;
    \item the competence proximity should be as low as possible when the coverage of a competence by all students is minimum.
\end{itemize}
For a competence $c \in C_p$ and a student $s$, we can visualise the above properties in the truth table in Table~\ref{tab:cp:truth:table}.
{
\begin{table}
    \centering
\begin{tabular}{c||c|c|c}
    $\mathrm{cvg} (\downarrow) \setminus w(\to) $ & $0$ & $(0,1)$ & $1$  \\\hline\hline
    $0$ & $1$& $1$ & $1$ \\\hline
    $(0,1)$ & $1$ & & $\sim \mathrm{cvg}$ \\\hline
    $1$ & $1$ & $\sim (1-w)$ & $0$
\end{tabular}
    \caption{\footnotesize Competence proximity truth table; $\mathrm{cvg}$ stands for $\cvg{c,C_s}$, and $w$ for $w_p(c)$.}
    \label{tab:cp:truth:table}
\end{table}   
}
If we think $\cvg{c,C_s}$ and $w_p(c)$, the importance of competence c in program p, as Boolean variables, we can interpret this table as a logical formula 
\[
 w_p(c) \Rightarrow \cvg{c,C_s} \equiv \big(1-w_p(c)\big) \vee \cvg{c,C_s}
\]

However, $\cvg{c,C_s}$ and $w_p(c)$ are continuous variables in $[0,1]$, so we model the `or' condition of the above logical formula as the `maximum' between the two variables.
As such, we define the competence proximity of a student for an internship program as:
\begin{definition}[{\em Student's Competence Proximity}]\label{def:scp}\em
Given a student $s \in S$, an internship program $p\in P$, and a competence assignment $\eta_p$,  the competence proximity of $s$ for $p$ with respect to $\eta_p$ is:
\begin{equation}
	\mathrm{cp} (s,p,\eta_p) = \prod_{c \in \eta_p(s)} \max \big\{ \big(1-w_p(c)\big), \cvg{c,C_s} \big\}.
\end{equation}
\end{definition}
Moving to the competence proximity of a team of students $K\subseteq S$ for program $p$, we use the Nash product of the competence proximity of the individuals in $K$ for $p$, with respect to some FCAF $\eta_p$.
The Nash product assigns a larger value to teams where all students equally contribute to their program, rather than to teams where some students have a small contribution.
\begin{definition}[{\em Team's Competence Proximity}]\label{def:team_comp_proximity}\em
Give a team $K$  a program $p\in P$, and a competence assignment $\eta_p$, the competence proximity of team $K$ for program $p$ is:
\begin{equation}
    \mathrm{cp}(K,p,\eta_p^K) = \prod_{s \in K} \mathrm{cp(s,p, \eta_p^K)}.
\end{equation}
\end{definition}

For a team $K$ and a program $p$ its competence proximity varies depending on the competence assignment at hand.
We define the best competence assignment as the {\em fair} one (Definition \ref{def:fair_competence_assignment}) that maximizes the competence proximity:
\begin{subequations}
\begin{align*}
\eta_p^{K\ *} &= \argmax_{\eta_p^K \in \Theta_p^K} \{\cp{K,p,\eta_p^K}\}\\
 &= \argmax_{\eta_p^K \in \Theta_p^K} \prod_{s \in K} \mathrm{cp(s,p, \eta_p^K)}\\
 &= \argmax_{\eta_p^K \in \Theta_p^K} \prod_{c \in \eta_p(s)} \max \big\{ \big(1-w_p(c)\big), \cvg{c,C_s} \big\}
\end{align*}
\end{subequations}
Finding the {\em best} competence assignment is optimization problem itself.
Even though the above is not a linear optimization problem, it can be easily linearized by considering the logarithm of $\cp\cdot$:
\begin{subequations}
\begin{align*}
    \eta_p^{K\ *} &= \argmax_{\eta_p^K \in \Theta_p^K} \{\cp{K,p,\eta_p^K}\}\equiv\argmax_{\eta_p^K \in \Theta_p^K} \{\log\{\cp{K,p,\eta_p^K}\}\}\\
    &=\argmax_{\eta_p^K \in \Theta_p^K}\log \Big\{ \prod_{s \in K} \mathrm{cp(s,p, \eta_p^K)} \Big\} = \argmax_{\eta_p^K \in \Theta_p^K}\sum_{s \in K} \log \big\{\mathrm{cp(s,p, \eta_p^K)}\big\}
\end{align*}
\end{subequations}

\subsection{The team allocation problem as an optimisation problem}\label{sec:sub:taip}

Finding a good allocation of students to a collection of internship programs is yet another optimization problem that tries to maximize the {\em overall} competence proximity of all teams for their assigned internship program.
That is, for a single program $p$, the best candidate team is the one that maximizes the competence proximity: $K^* = \argmax_{K \in \mathcal{K}_p}\ \mathrm{cp}(K,p)$.
$K^*$ is the best candidate when a single program is at hand.
For a set of programs $P$, with $|P|>1$, we need to maximize the competence proximity of all candidate teams with their corresponding programs.
Suppose we have a team assignment function $g:P \to 2^S$, which maps each $p\in P$ with a team of students $K \in \mathcal{K}_p$.
We assume that for two programs $p_1$ and $p_2$ it holds that $p_1 = p_2 \Leftrightarrow g(p_1) = g(p_2)$.
In the setting of matching internship programs with teams of students we should consider only team assignment functions $g$ such that $g$ assigns each student to at most one program.
As such, we can define {\em feasible} team assignment functions:
\begin{definition}[{\em Feasible Team Assignment Functions (FTAF)}]\em
Given a set of programs $P$ and a set of students $S$, a feasible team assignment function $g \in G$ is such that for each pair of programs $p_1, p_2 \in P$ with $p_1 \neq p_2$, it holds that $g(p_1) \cap g(p_2) = \emptyset$; and for all $p \in P$ it holds that $|g(p)| = m_p$.
\end{definition}
The family of all feasible team assignments is denoted with $G_{\mathrm{feasible}}$. Now we are ready to formalise our team allocation problem as follows:

\begin{definition}[Team Allocation for Internship Programs Problem (TAIPP)]
\label{def:TAIPP}
Give a set of internship programs $P$, and a set of students $S$, the team allocation for internship programs problem is to select the team assignment function $g^* \in G$ that maximizes the overall competence proximity: 
    \begin{equation}\label{eq:formal:optimization:problem}
        g^* = \argmax_{g \in G_{feasible}}\ \prod_{p\in P}\cp{ g(p),p,\eta_p^{g(p)\ *}}
    \end{equation}
\end{definition}

The following result establishes that the TAIPP is $\np-complete$ by reduction to a well-known problem in the MAS literature.

\begin{theorem}
The TAIPP, with more than one program at hand, is $\np-complete$.
\end{theorem}
\begin{proof}
The problem is in $\np$ since we can decide whether a given solution is feasible in polynomial time ($\mathcal{O}(\sum_{p \in P} m_p)$).
We show that the problem is $\np-complete$ by using a reduction from {\em Single Unit Auctions with XOR Constraints and Free Disposals} (referred to as BCAWDP with XOR Constraints) which is shown to be $\np-complete$~\cite{10.1145/544741.544760}.
In the BCAWDP with XOR Constraints, the auctioneer has $N$ items to sell, the bidders place their bids $B_i = \langle \mathbf{b}_i,b_i\rangle$ with $\mathbf{b}_i$ be a subset of items and $b_i$ the price. Between two bids can exist an XOR constraint--not necessarily to every pair of bids. The auctioneer allows free disposals, i.e., items can remain unsold.
Given an instance of BCAWDP with XOR Constraints, we construct an instance of student-teams allocation to internship programs problem as follows: ``For each item $i$ we create a student $s_i$. For each program $p_j$ of size $m_{p_j}$ we create $\binom{|S|}{m_{p_j}}$ different bids $B_{jk} = \langle \mathbf{b}_{jk},b_{jk} \rangle$, where $|S|$ is the number of items, $|\mathbf{b}_{jk}| = m_{p_i}$, and $b_{jk} = \cp{\mathbf{b}_{jk},p_j, \eta_{p_j}^{\mathbf{b}_{jk} \ *} }$.
All bids created for program $p_j$ are XOR-constrained bids.
Moreover, each pair of bids $B_{j,k},B_{q,l}$ such that $\mathbf{b}_{jk} \cap \mathbf{b}_{ql} \neq \emptyset$ are also XOR-constrained.''
Now the team allocation for internship programs problem has a feasible solution if and only if BCAWDP with XOR constraints has a solution.
\end{proof}

Typically, the winner determination problem for combinatorial auctions can be cast and solved as a linear program. Along the same lines, we propose how to solve the TAIPP by meas of LP in Sec~\ref{sec:optimization:problem}. Before that, the following section characterises the search space with which an algorithm solving the TAIPP must cope. 

\subsection{Characterising the search space}

The purpose of this section is to characterise the search space defined by the TAIPP. This amounts to quantifying the number of feasible team assignment functions in $G_\mathrm{feasible}$.
For that, 
we start by splitting the programs in $P$ into $k$ buckets of programs, where the programs in the same bucket require teams of the same size.
That is,  we have $b_1,\cdots,b_k \subseteq P$ buckets where $b_i \cap b_j = \emptyset,\ \forall i,j=1,\cdots,k$ and $\bigcup_{i=1}^k b_i = P$. For each bucket $b_i$ with $|b_i|=n_i$, it holds that $m_{p_1} = m_{p_2} = \cdots = m_{p_{n_i}} = m_i$ for all $p_1,p_2,\cdots, p_{n_i} \in b_i$; and $m_i\neq m_j$, that characterise $b_i$ and $b_j$ respectively, for any $i\neq j = 1, \cdots, k$. Next, we will distinguish three cases when counting the number of feasible teams in $G_\mathrm{feasible}$:
\begin{itemize}
    \item[-] Case I  : $\sum_{p\in P} m_p = \sum_{i=1}^k m_i\cdot |b_i| = N$, we have exactly as many students as required by all programs in $P$. 
    In this case, we seek for partition functions over $P$. The space of $G_\text{feasible}$ is $\frac{N!}{ \prod_{i=1}^k (m_i!)^{b_i}}$ according to Theorem 3.4.19 in~\cite{Maddox:2002:isb0124649769}.
    \item[-] Case II : $\sum_{p\in P} m_p = \sum_{i=1}^k m_i\cdot |b_i| < N$, we have more students than the required ones by all programs in $P$.
    Following the Example 3.4.20 in~\cite{Maddox:2002:isb0124649769}, we assume one more bucket $b_{k+1}$ containing exactly one {\em auxiliary} program, which requires a team of size $m_{k+1} = \sum_{i=1}^k m_i\cdot |b_i| - N$.
    Now there are $|G_\mathrm{feasible}| = \frac{N!}{\prod_{i=1}^k (m_i!)^{b_i} \cdot \big( N- \sum_{i=1}^k |b_i|\cdot m_i\big)!}$ different feasible team assignment functions.
    \item[-] Case III: $\sum_{p\in P} m_p = \sum_{i=1}^k m_i\cdot |b_i| > N$, we have less students than the required ones by all programs in $P$. 
    In this case, first we need to introduce $cover(P,S) = \{P' \subset P : \sum_{p \in P'} m_p \leq N \wedge \not\exists\ p' \in P-P' : m_{p'} \leq N-\sum_{p \in P'} m_p\}$ as the set that contains all the subsets of programs $P'\subset P$ such that $S,P'$ leads to Case I or Case II, and by adding any $p\not \in P'$ in $p'$ it will lead to Case III.
    The number of feasible team assignment functions is:
    \[|G_\mathrm{feasible}| = \sum_{P'\in cover(P,S)}\frac{N!}{\prod_{i=1}^k (m_i!)^{b_i} \cdot \big( N- \sum_{i=1}^k |b_i|\cdot m_i\big)!}\]
    where variables $k,b_1,\cdots, b_k$ and $m_1,\cdots, m_k$ changes according to $P'$.
    The size of set $cover(P,S)$ depends on the total number of students, and the team sizes required by the programs in $P$.\label{itemize:hard_case}
\end{itemize}

Note that the number of feasible team assignment functions quickly grows with the number of programs and students, hence leading to very large search spaces.

\section{Solving The TAIPP as a linear program}\label{sec:optimization:problem}

In what follows we show how to solve the TAIPP in Definition \ref{def:TAIPP} as an LP. First, for each time $K \subseteq S$ and program $p  \in P$, we will consider a binary decision variable $x_K^p$. The value of $x_K^p$ indicates whether team $K$ is assigned to program $p$ or not as part of the optimal solution
of the TAIPP. Then, solving the TAIPP amounts to solving the following non-linear program:

\begin{equation}
    \max \prod_{p \in P} \prod_{k \in \mathcal{K}_p} \Big(\cp{K, p,\eta_p^{K\ *}}\Big)^{x_K^p} \label{eq:nlp}
\end{equation}
subject to: 
\begin{subequations}
\begin{align}
&\sum_{K \subseteq S} x_K^p \cdot \mathbbm{1}_{K \in \mathcal{K}_p} \leq 1 & \forall p \in P \label{constr:at_most_one_team}\\
    &\sum_{p\in P} \sum_{K \subseteq S} x_K^p \cdot \mathbbm{1}_{s \in K} \cdot \mathbbm{1}_{K \in \mathcal{K}_p} \leq 1 & \forall s \in S \label{constr:at_most_one_program}\\
    &x_K^p \in \{0,1\}&\forall K\subseteq S, p \in P
    \label{constr:domain}
    \end{align}
    \label{eq:csp++}
\end{subequations}

Constraint \ref{constr:at_most_one_team} ensures that a program is allocated a single team. Constraint \ref{constr:at_most_one_program} ensures that any two teams sharing some student cannot be assigned to programs at the same time. Notice that the objective function (see Eq~\ref{eq:nlp}) is non-linear. Nevertheless, it is easy to linearise it by maximising the logarithm of \linebreak[4] $\prod_{p \in P} \prod_{k \in \mathcal{K}_p} \Big(\cp{K, p,\eta_p^{K\ *}}\Big)^{x_K^p}$. Thus, solving the non-linear  program above is equivalent to solving the following binary linear program:
\begin{equation}
\operatorname{max} \sum_{p \in P} \sum_{K \in \mathcal{K}_p}  {x_K^p} \cdot \log\Big(1 + \cp{K, p,\eta_p^{K\ *}}\Big)
\label{eq:lp}
\end{equation}
subject to: equations \ref{constr:at_most_one_team}, \ref{constr:at_most_one_program}, and \ref{constr:domain}. 
Therefore, we can solve this LP and solve with the aid of an off-the-shelf LP solver such as, for example, CPLEX, Gurobi, or GLPK. If given sufficient time, an LP solver will return an optimal solution to the TAIPP.

At this point, it is worth mentioning that computing the objective function in \ref{eq:lp} to build the LP requires the pre-computation of the values of $\cp{K, p,\eta_p^{K\ *}}$, which amounts to solving an optimisation problem per each pair of team and program. This is bound to lead to large linear programs as the number of students and programs grow. Furthermore, an LP solver is a general-purpose solver that does not exploit the structure of the problem. Thus, in the next section we introduce the {\em TAIP} algorithm, an anytime algorithm based on local search that yields approximate solutions to the TAIPP. Unlike an LP solver, TAIPP is a specialised algorithm does exploit the structure of TAIPP instances. Section \ref{sec:experiments} will show that TAIPP manages to outperform a general-purpose LP solver.

\section{A heuristic algorithm for TAIPP}\label{sec:TAIP}
 The TAIP algorithm consists of two stages: {\em (a)} finding an initial feasible allocation of students to programs, and {\em (b)} continuously improving the best allocation at hand by means of swaps between team members.
 
\vspace{-10pt}
\subsection{Initial team allocation}
\label{subsec:initialAllocation}

During this stage the algorithm finds an initial feasible team allocation.
The algorithm sequentially picks a team for each program, starting from the `hardest' program to the `simplest' one.
Intuitively, `hard' programs are more selective, i.e., there are a few students that can cover it; as such, picking teams for the harder programs first is easier as we have more options (students) available.
In order to evaluate the hardness of a program we will be using the notion of fuzzy entropy.

\begin{figure}
\includegraphics[width=0.35\textwidth]{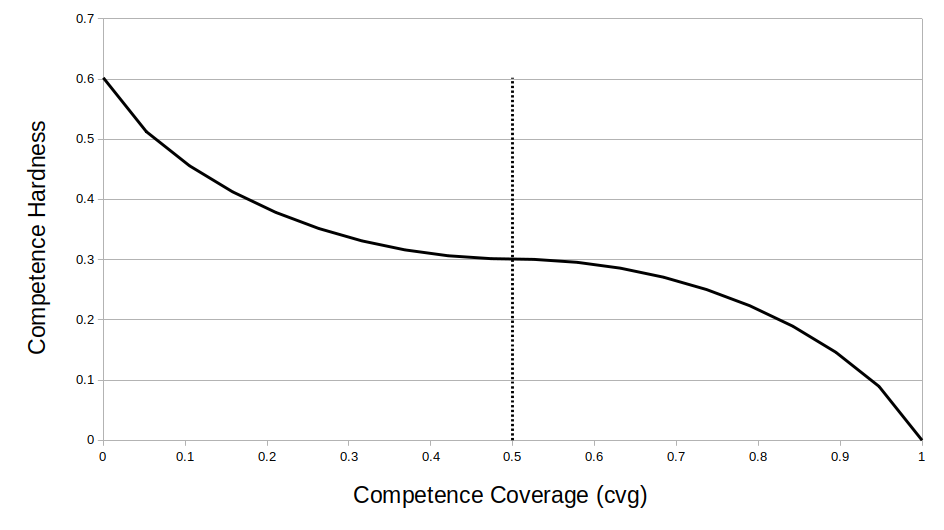}
\caption{\footnotesize Competence hardness.}
\label{fig:competence:hardness}
\end{figure}

To begin with, we first evaluate the required competences from all programs, as to how hard is for the students to cover them.
Looking at the competence coverage metric, we can view it as a {\em membership function}~\cite{ZADEH1965338}, i.e., a function that indicates in what degree a competence lies in a set of competences.
Thus, fuzzy entropy~\cite{DELUCA197455,1008855} indicates the difficulty of finding students to cover a competence.
However we need to discern two extreme cases:
\begin{itemize}
    \item[-] all students possess competence $c$, i.e. $\cvg{c,C_s}=1\ \forall s\in S$;
    \item[-] no student can cover competence $c$, i.e. $\cvg{c,C_s}=0\ \forall s\in S$.
\end{itemize}
Although, the above two cases result with the same fuzzy entropy ($0$), their intuitive interpretation is exactly the opposite.
In the former case, finding a student for covering this competence within a team it is trivial since everyone can cover it.
In the latter case, finding a student for covering this competence within a team it is trivial since no-one can cover it.
Thus, in our definition of competence hardness we exploit the notion of fuzzy entropy, but we also embrace the intuitive interpretations above. Formally: 

\begin{definition}[Competence Hardness]
Given a set of students $S$, the hardness of a competence $c$ is defined as
\begin{equation}
h(c,S) = -K \sum_{s \in S} \mathcal{H}\big(\cvg{c,C_s} \big)
\end{equation}
where $K=1/|S|$ is a normalization factor,\\$
    \mathcal{H}(x) = \begin{cases}  H(x) + H(1-x)& \text{if } x\geq0.5\\
    4 \cdot H(0.5) - H(x) - H(1-x) & \text{otherwise }
    \end{cases}$,\\
    and $H(x) = x\cdot\log(x)$.
\end{definition}
The hardness of a competence $c$ coincides with its fuzzy entropy when for all students the competence coverage is greater than $0.5$.
If for all students the competence coverage is less than $0.5$  the competence hardness is the constant $4\cdot 0.5 \cdot \log(0.5)$ minus the fuzzy entropy.
The constant $4\cdot 0.5 \cdot \log(0.5)$ derives from the fuzzy entropy of point $0.5$: coverage $0.5$ indicates that all students are neither good nor bad for this competence, as such hardness in point $0.5$ shall be the median, thus the maximum of the competence hardness is $2\cdot \big(0.5\log(0.5) +(1-0.5)\log(1-0.5)\big) = 2 \cdot 2 \cdot 0.5\log(0.5)$. Graphically, the competence hardness is shown in Fig~\ref{fig:competence:hardness}.

The degree of hardness of a program is determined by the available set of students' difficulty for covering the program's competencies.
We remind the reader that each required competence is accompanied by an importance weight (Sec~\ref{sec:basic:elements}).
Thus, consider a program where its most important competence $c_{important}$ (i.e., the competence with the highest $w_p(c)$) is very difficult to be covered $h(c_{important},S) \simeq 4\cdot H(0.5)$, then this program is extremely hard.
On the other hand, if a specific competence $c$ is somewhat difficult to be covered ($h(c_{important},S) \to 4\cdot H(0.5) $), but it is not very important ($w_p(c)\to 0$), then the program is not that hard.

\begin{definition}[Program Hardness]
Given a set of students, the hardness of a program $p$ is defined as the aggregation of the hardness for the competences in the program weighted by the importance of each competence: $ h(p,S) = W\cdot \sum_{c \in C_p}  \frac{1}{h(c,S) + \epsilon} \cdot {w_p(c)}$, where $W = \frac{1}{\sum_{c\in C_p} w_p(c)}$ is a normalization factor, and $\epsilon$ is  a small positive constant.
\end{definition}
In words, the more important and the more difficult a competence is to be covered, the more difficult is to find students with high competence proximity for the program, consequently the harder the program is considered to be.
Note that both $w_p(c)$ and $h(c)$ are non-negative, so the hardness of one competence cannot be counteracted by the non-hardness of another within the same program.

\begin{algorithm}
\caption{Initial Team Allocation}
\label{alg:ita}
\Input{Students $S$, programs $P$, (optionally) sorting order $order$ for $P$}
\Output{team assignment function $g$}
$V_p \leftarrow \bigcup_{p \in P} C_p$\;
\lFor{$c \in V_p$}{$\mathrm{hc}[c] \leftarrow h(c,S)$}
\lFor{$p \in P$}{$\mathrm{hp}[p] \leftarrow h(p,S)$}
sort $P$ in descending order {\em wrt} $\mathrm{hp}$\;
\While{$P \neq \emptyset$}{
    $p \leftarrow$ pop first from $P$\;
    \If{$|S|\geq m_p$ and $\mathrm{hp}[p]<1$}{
    sort $S$ maximizing coverage in benches of $|C_p|$\;
    \tcc{assign team to program}
    $g(p) \leftarrow m_p$ fist students in $S$\;
    $S \leftarrow S \setminus g(p)$\;
    update values in $\mathrm{hc}$\;
    \lIf{$|S|<1$}{break}
    }
}
\Return $g$\;
\end{algorithm}
\normalsize


\subsection{Improving team allocation}

In the second stage we perform a number of random `movements', until convergence to a local or global maximum.
The second stage starts with the team assignment produced in the first stage.
Thereafter, we iteratively improve the  current team assignment either {\em (i)} by employing crossovers of students between two programs, and/or {\em (ii)} by swapping assigned students with available ones if they exist.
Specifically, following Algorithm~\ref{alg:imta} within an iteration we randomly pick two programs~(line~\ref{alg:imta:pick:projects}) and attempt to improve the competence proximity of the pair by exhaustively searching of all possible crossovers of the students assigned to these programs~(line~\ref{alg:imta:exhaustive:search}).
However, in order not to computationally overload our algorithm with repetitive exhaustive searches, we perform it only if the following two conditions hold:
\begin{enumerate}
	\item the two programs share similar competencies; and
	\item sine student in one of the teams improves the coverage of some competence of the other team.
\end{enumerate}
In order to evaluate if two programs share similar competencies we exploit the {\em Hausdorff} distance~\cite{conci2018distance}.
The Hausdorff distance between the required competencies of two programs $p_k$ and $p_l$ is defined as: 
\[\mathrm{dist} (C_{p_1},C_{p_2}) = \max \big\{ \min_{c \in C_{p_1}}\{\cvg{c,C_{p_2}} \},\min_{c \in C_{p_2}} \{ \cvg{c,C_{p_1}} \} \big\}.\]
The above two conditions encode the potentiality of finding an improvement for these two programs, and whether it is worth performimg an exhaustive search.
In the exhaustive search, given the students $g(p_k)\cup g(p_l)$ we produce all possible partitions that contain two teams of sizes $m_{p_k}$ and $m_{p_l}$. For each of these partitions we compute the competence proximity of the pair of programs, and yield with the optimum one, i.e., with the partition that achieves the greater competence proximity.

In case we did not achieved any improvements from the previous step and there are available students, i.e. student that have not been assigned to any program, we attempt to swap assigned students with available ones~(line ~\ref{alg:imta:local:swaps}).
That is, we randomly pick a student from either of the two programs, and try to randomly swap them with a student in $S_{available}$. 
If we achieve an improvement we keep this alteration, otherwise we repeat this process for a fixed number of attempts.
In case we succeeded to imporove the competence proximity of the pair, we update the team assignement $g$, the set of availble students $S_{available}$, and the current overall competence proximity (lines~\ref{alg:imta:update:g}-\ref{alg:imta:update:g:end}).

In order to overcome the possibility of a series of unsuccessful attempts between random programs, we force a more `systematic' search, which we call local search, on the programs. This local search is performed after a constant number of iterations~(line~\ref{alg:imta:local:search}).
In the local search (line~\ref{alg:imta:local:search2}) we go through {\em all} programs in $P$, swap all members, and check whether some swap improves the overall competence proximity--in the swaps we consider all students: both assigned and available.

Note that Algorithm~\ref{alg:imta} is anytime algorithm that can yield a result after any number of iterations indicated by the user.
However, in its generality, we adopt a notion of convergence in order to terminate the algorithm.
That is, we terminate the algorithm (line~\ref{alg:imta:while:outer}):
\begin{itemize}
 	\item[-] after a number of iterations without no improvements; {\em or}
 	\item[-] if we reach an overall competence proximity close to $1$.
\end{itemize} 
 Note that we added the latter termination condition in order to avoid unnecessary iterations until convergence, due to the fact that the maximum value the overall competence proximity can reach is $1$.
 We remind the reader that the competence proximity is the Nash product of the individual competence proximity of the teams to their assigned program (Eq~\ref{eq:nlp}), and each individual competence proximity lies in $[0,1]$ (Def~\ref{def:team_comp_proximity}).
 However, we should make clear that the overall competence proximity does {\em not always} reach $1$, but that it can {\em never exceed} $1$.
\begin{algorithm}
\caption{Improve Team Allocation}
\label{alg:imta}
\Input{Students $S$, programs $P$, team assignment $g$}
\Output{improved team assignment $g$}
    $S_\mathrm{available} = S \setminus \bigcup_{p \in P} g(p)$\;
    $\mathrm{current\_cp} = \prod_{p \in P} \cp{g(p),p}$\;
    \While{non\_improved {\bf and} $1-\mathrm{current\_cp}>\varepsilon$}{\label{alg:imta:while:outer}
        $p_k, p_l \leftarrow$ randomly select two programs from P\;\label{alg:imta:pick:projects}
        
        $\mathrm{pair\_cp} = \cp{g(p_k),p_k} \cdot \cp{g(p_l),p_l}$ \;\label{alg:imta:compute:pair:cp}
        \If{\em potentiality($p_1,p_2,g$)}{\label{alg:imta:check:potentiality}
            $\mathrm{new\_cp},K_k,K_l \leftarrow$ exhaustiveSearch($p_1,p_2,g$)\;\label{alg:imta:exhaustive:search}
        }\Else{
            $\mathrm{new\_cp},K_k,K_l \leftarrow$ localSwaps($p_1,p_2,g,S_\mathrm{available}$)\;\label{alg:imta:local:swaps}
        }\label{alg:imta:check:potentiality:end}
        \If{$\mathrm{new\_cp} > \mathrm{pair\_cp}$}{\label{alg:imta:update:g}
            $g(p_k) \leftarrow K_k$\;
            $g(p_l) \leftarrow K_l$\;
            $S_\mathrm{available} \leftarrow S \setminus \bigcup_{p \in P} g(p)$\;
            $\mathrm{current\_cp} \leftarrow \mathrm{current\_cp}\cdot \frac{\mathrm{new\_cp}}{\mathrm{pair\_cp}}$\;
            $\mathrm{pair\_cp} \leftarrow \mathrm{new\_cp}$\;
        }\label{alg:imta:update:g:end}
        \If{time for local search}{\label{alg:imta:local:search}
            $g,S_\mathrm{available},\mathrm{current\_cp}\leftarrow$ localSearch($P,g,S_\mathrm{available}$)\;\label{alg:imta:local:search2}
        }\label{alg:imta:local:search:end}
    }\label{alg:imta:while:outer:end}
    \Return $g$\;
\end{algorithm}

\normalsize

\section{Empirical analysis}\label{sec:experiments}

The purpose of this section is to empirically evaluate the TAIP algorithm along four directions:
\begin{itemize}
    \item the quality of the solutions that it produces in terms of optimality;
    \item the quality of the solutions produced by the initial stage;
    \item the time required by TAIP to produce optimal solutions with respect to CPLEX, an off-the-shelf linear programming solver; and
    \item the time required by TAIP to yield optimal solutions as the number of students and programs grow.
\end{itemize}

Overall, our results indicate that TAIP significantly outperforms CPLEX, and hence it is the algorithm of choice to solve the Team Allocation for Internship Programs Problem introduced in this paper. Next, in Sec \ref{subsec:settings} we describe the settings employed in our experiments, whereas Sec \ref{subsec:results} dissects our results. 

\subsection{Empirical settings}
\label{subsec:settings}
For our experimental evaluation we used an existing competence ontology provided by {\em Fondazione Bruno Kessler} (https://www.fbk.eu/en/); and generated synthetic data  in the following way:
\paragraph{Internship program generation} For each program $p$
\begin{enumerate}
    \item select the required team size $m_p\sim \mathcal{U}\{1,3\}$
    \item select the number of required competences $|C_p| \sim \mathcal{U}\{2,5\}$
    \item randomly choose $|C_p|$ competences from the ontology
    \item the required level function is set to $l_p(c) = 1,\ \forall\ c\in C_p$
    \item the weight function is $w_p(c) = \mathcal{N}\big(\mu=\mathcal{U}(0,1),\sigma=\mathcal{U}(0.01,0.1) \big)$ bounded in $(0,1]$ for all $c \in C_p$.
\end{enumerate}
\paragraph{Student generation} For each program $p$
\begin{enumerate}
    \item generate $m_p$ new students such that for each student $s$:
    there are competences $c\in C_p$ and $c'\in C_s$ such that $c'$ is
    {\em (i)} identical to $c$; or {\em (ii)} a child-node of $c$ in the ontology;
    uniformly selected among the options.
\end{enumerate}

With these generators we constructed 60 different TAIPP instances, which are shown in Table~\ref{tab:datasets}. We solve each problem instance with both TAIP and the IBM CPlex linear programming (LP) solver. The experiments were performed on a PC with Intel Core i7 (8th Gen)  CPU, 8 cores, and 8Gib RAM. Moreover, we employed IBM ILOG CPLEX V12.10.0.
For all implementations we used Python3.7.

\subsection{Results}
\label{subsec:results}

\begin{figure*}
    \centering
    \includegraphics[width=0.6\textwidth]{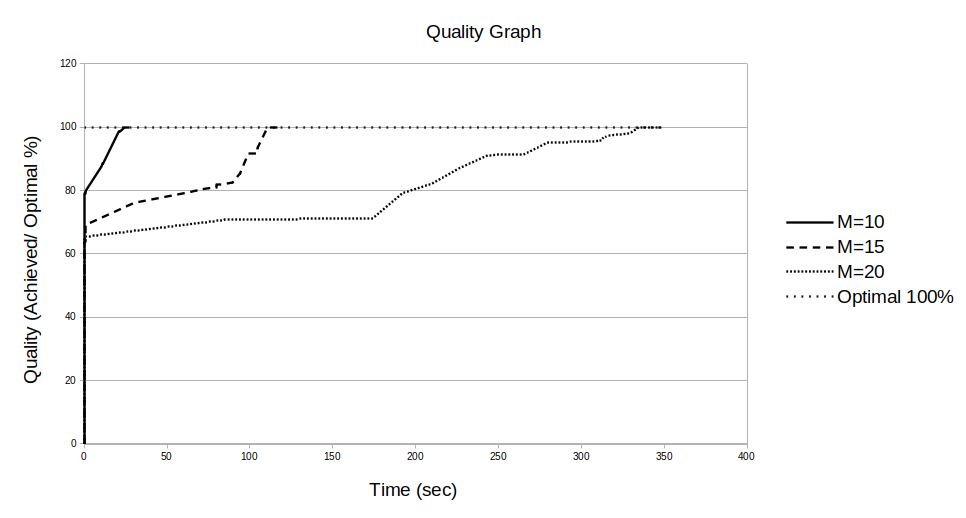}
    \caption{\footnotesize Solution quality achieved by TAIP along time.}
    \label{fig:quality}
\end{figure*}

\noindent
\textbf{Quality analysis.}
Using the optimal solutions yielded by CPLEX as an anchor, we can evaluate the quality of the solutions computed by the TAIP algorithm. Notice that for all problem instances, TAIP reaches the optimal solution.
More precisely, for every problem instance, TAIP achieved a solution whose value, in terms of competence proximity, is the same as the value of the optimal solution computed by CPLEX.
Fig~\ref{fig:quality} shows the average {\em quality ratio} of TAIPP with respect to CPLEX along time for the problem instances in Table \ref{tab:datasets}.
We calculate the quality ratio by dividing the competence proximity computed by TAIP by the optimal value computed by CPLEX, and it is depicted as a percentage (\%).

\begin{figure*}
\centering
\begin{subfigure}[b]{0.6\textwidth}
    \includegraphics[width=\textwidth]{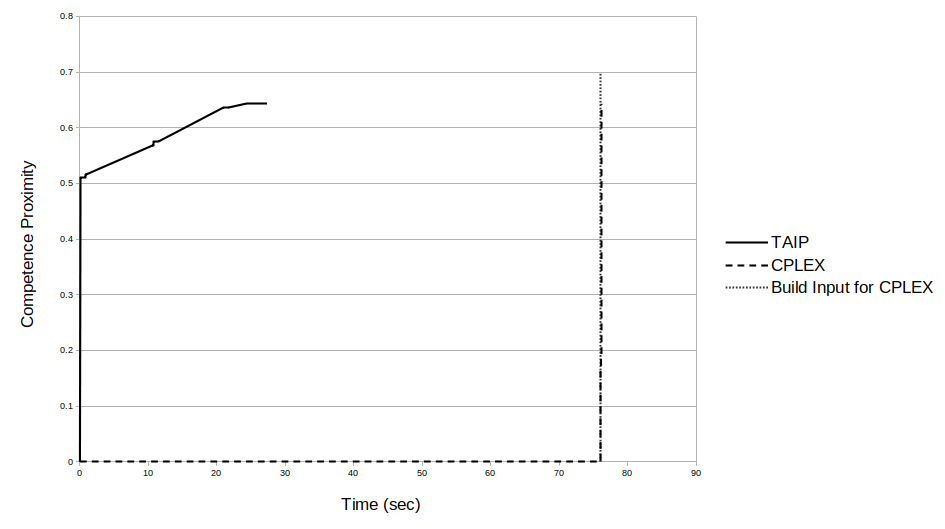}
    \caption{\footnotesize 10 programs}
    \label{fig:TAIPvsCPEX:M10}
\end{subfigure}
\begin{subfigure}[b]{0.60\textwidth}
    \includegraphics[width=\textwidth]{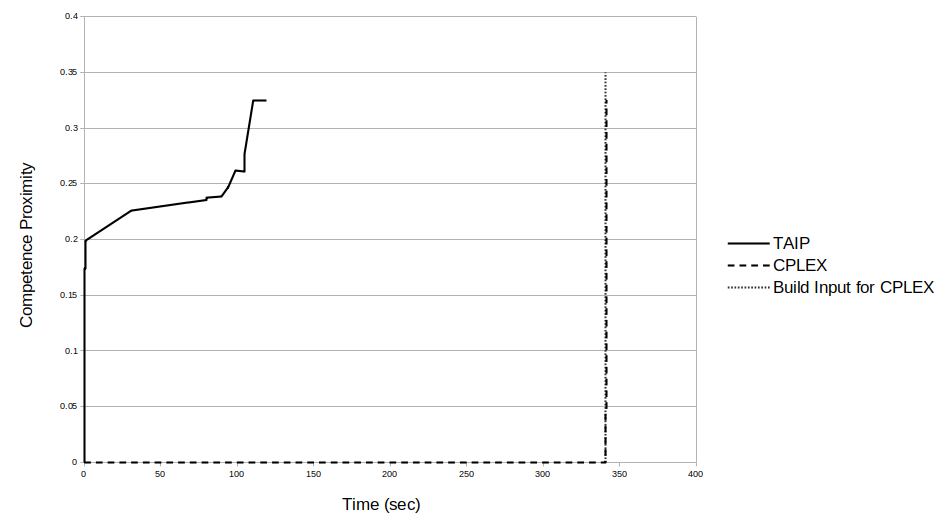}
    \caption{\footnotesize 15 programs}
    \label{fig:TAIPvsCPEX:M15}
\end{subfigure}
\begin{subfigure}[b]{0.60\textwidth}
    \includegraphics[width=\textwidth]{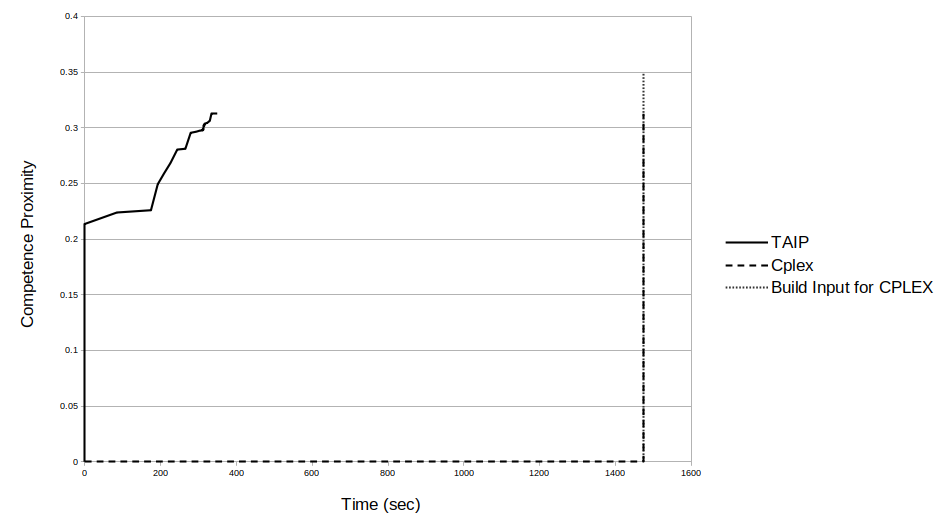}
    \caption{\footnotesize 20 programs}\label{fig:TAIPvsCPEX:M20}
\end{subfigure}
\caption{\footnotesize Average Competence Proximity vs Time}
\label{fig:TAIPvsCPEX}
\end{figure*}

\noindent
\textbf{Runtime analysis.}
The greatest advantage of TAIP is that it is way much faster than CPLEX.
As shown in Fig~\ref{fig:TAIPvsCPEX} TAIP reaches optimality in less than half of the time required by CPLEX.
Specifically, for problem instances with 10 programs, TAIP requires on average $\sim40\%$ of the time CPLEX needs, i.e., is $\sim60\%$ faster.
As to problem instances with 15 programs, TAIP requires on average $\sim45\%$ of the time employed by CPLEX ($\sim55\%$ faster). Finally, for problem instances with 20 programs, TAIP requires on average $\sim29\%$ of the time spent by CPLEX ($\sim71\%$ faster). Therefore, the larger the size of the problem instances, the larger the benefits for TAIP with respect to CPLEX.
Here we should note that the time consuming task for CPLEX is the building of the LP encoding the problem, while solving the actual problem is done in seconds.
This indicates that the problem instances under investigation are rather large than hard: as the number of programs increases, so does the number of students, resulting in large linear programs.



\begin{figure*}
\centering
\begin{subfigure}[b]{0.60\textwidth}
   \includegraphics[width=\textwidth]{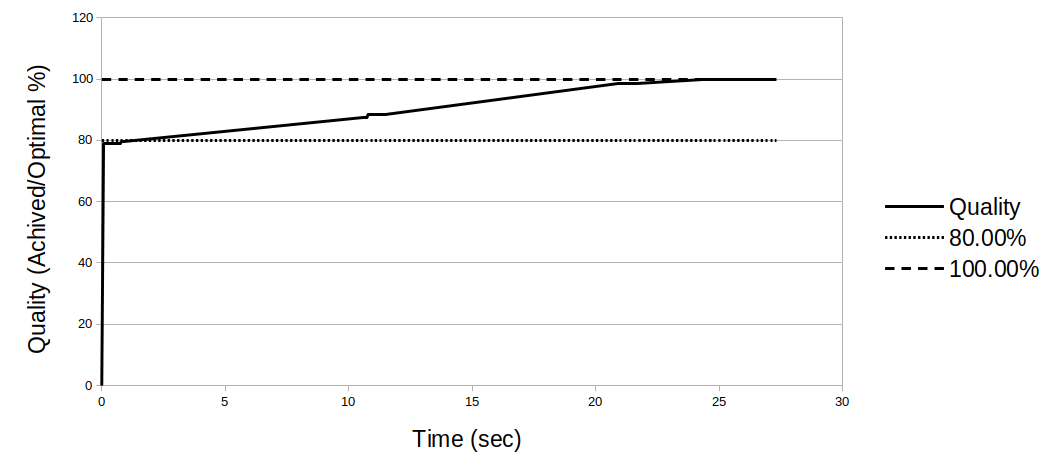}
    \caption{\footnotesize 10 programs}
    \label{fig:TAIP:M10}
\end{subfigure}
\begin{subfigure}[b]{0.60\textwidth}
    \includegraphics[width=\textwidth]{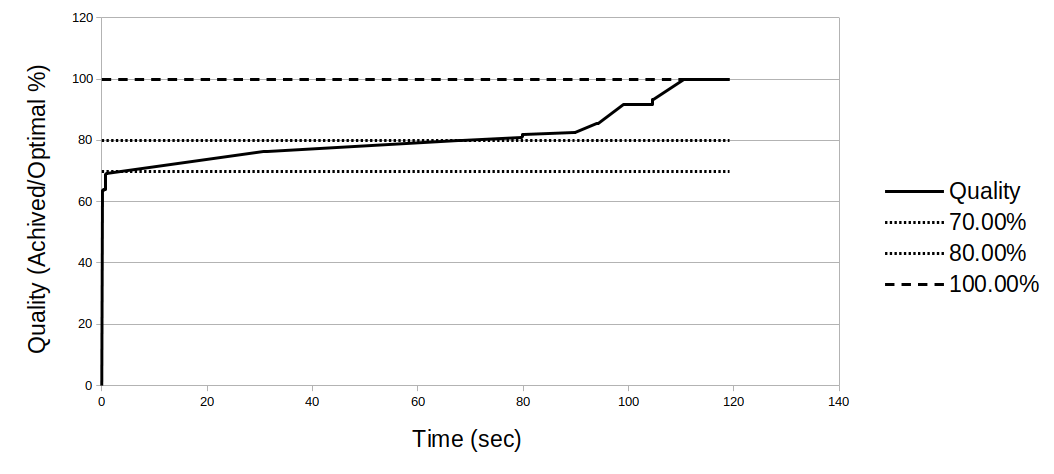}
    \caption{\footnotesize 15 programs}
    \label{fig:TAIP:M15}
\end{subfigure}
\begin{subfigure}[b]{0.60\textwidth}
     \includegraphics[width=\textwidth]{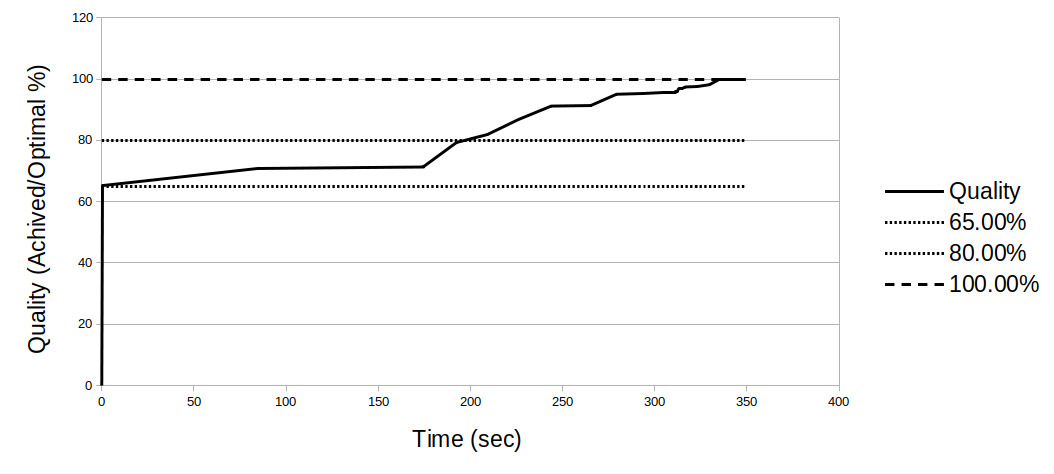}
    \caption{\footnotesize 20 programs}
    \label{fig:TAIP:M20}
\end{subfigure}
\caption{\footnotesize Anytime Behavior}
    \label{fig:TAIP}
\end{figure*}

\noindent
\textbf{Anytime analysis.}
Last but not least we present our results on the anytime behavior of TAIP, as shown in Fig~\ref{fig:TAIP}. We observe that  after completing the initial stage described in Sec~\ref{subsec:initialAllocation}, the solution quality produced by TAIP reaches 80\%, 70\%, and 65\% of the optimal solution, for problem instances with 10,15 and 20 programs respectively.
Furthermore, TAIP reaches quality 80\% in $0.001 \times t_{CPLEX}$ for 10 programs, 70\% in $0.025\times t_{CPLEX}$ for 15 programs, and 65\% in $0.0002 \times t_{CPLEX}$ for 20 programs, where $t_{CPLEX}$ is the time CPLEX needs to compute the optimal solution.
Moreover, in all investigated settings we reached 80\% quality in less than 20\% of the time CPLEX needs: 0.1\%, 20\%, and 13.5\% of CPLEX time for 10,15, and 20 programs.



\section{Conclusions and future work}

Here we formally defined the problem of {\em Team Allocation for Internship Programs Problem}. We first studied the problem's complexity and characterised its search space.
Thereafter, we provided an encoding to otimally solve the TAIPP by means of linear programming.
Then, we proposed a novel, heuristic anytime algorithm, TAIP.
Finally, we conducted a systematic comparison of TAIP versus the CPLEX LP solver when solving TAIPP problem instances. Our experimental evaluation showed that TAIP outperforms CPLEX in time, mainly because of the extremely large input that the latter requires.
Moreover, TAIP always managed to reach the optimal solution for the problem instances under investigation.
Specifically TAIP converged to the optimal in less than 40\% of the time required by CPLEX, and achieved a quality of 80\% in less than 20\% of the time required by CPLEX.
As future work, we itend to device more intelligent strategies during the second state of TAIPP, instead of our current randomized strategy.
Furthermore, in the future we will study the performance of TAIP on actual-world data.
\begin{table*}
    \begin{subtable}{\textwidth}
    \centering{
    \begin{tabular}{c|c|c|c|c|c|c|c|c|c|c|c|c|c|c|c|c|c|c|c|c||c}
       Dataset & 1& 2& 3& 4& 5& 6& 7& 8& 9& 10& 11& 12& 13& 14& 15& 16& 17& 18& 19&20 & Average\\\hline
       N=\#Students&18  & 20  & 21  & 19  & 22  & 19  & 24  & 18  & 19  & 20  & 23  & 20  & 18  & 19  & 25  & 21  & 25  & 20  & 17  & 13 & 20.5
    \end{tabular}
    \caption{\footnotesize Family of datasets with 10 programs}\label{tab:family10}
    }
    \end{subtable}
    
    \begin{subtable}{\textwidth}
    \centering{
    \begin{tabular}{c|c|c|c|c|c|c|c|c|c|c|c|c|c|c|c|c|c|c|c|c||c}
       Dataset & 21& 22& 23& 24& 25& 26& 27& 28& 29& 30& 31& 32& 33& 34& 35& 36& 37& 38& 39&40 & Average\\\hline
       N=\#Students & 23  & 33  & 32  & 29  & 33  & 40  & 31  & 32  & 28  & 25 & 27  & 31  & 29  & 28  & 32  & 29  & 29  & 32  & 34  & 29& 30.6
    \end{tabular}
    \caption{\footnotesize Family of datasets with 15 programs}\label{tab:family15}
    }
    \end{subtable}
    
    \begin{subtable}{\textwidth}
    \centering{
    \begin{tabular}{c|c|c|c|c|c|c|c|c|c|c|c|c|c|c|c|c|c|c|c|c||c}
       Dataset & 41& 42& 43& 44& 45& 46& 47& 48& 49& 50& 51& 52& 53& 54& 55& 56& 57& 58& 59&60 & Average\\\hline
       N=\#Students  &  44  & 45  & 44  & 45  & 38  & 42  & 42  & 47  & 41  & 44 & 37  & 36  & 42  & 37  & 47  & 32  & 40  & 37  & 44  & 43 & 41.35
    \end{tabular}
    \caption{\footnotesize Family of datasets with 20 programs}\label{tab:family20}
    }
    \end{subtable}
    \caption{\footnotesize Synthetic problem instances.}
    \label{tab:datasets}
\end{table*}

\bibliographystyle{ACM-Reference-Format}  
\bibliography{sample-bibliography}  

\clearpage

\end{document}